\RequirePackage{fix-cm}
\documentclass[twocolumn]{svjour3}          
\smartqed  
\usepackage{amsmath}
\usepackage{graphicx}
\usepackage{multirow,multicol}
\usepackage{booktabs}
\usepackage{array}
\usepackage{tabularx}
\usepackage{CJKutf8}
\usepackage{amssymb}
\usepackage{xcolor}
\usepackage{stfloats}
\usepackage{natbib}
\setcitestyle{aysep={}} 
\usepackage{natbibspacing}
\usepackage{etoolbox}
\usepackage[misc]{ifsym}
\usepackage{tabto}
\usepackage[hyperindex,pdftex]{hyperref}
\usepackage{comment}
\usepackage[ruled,lined,boxed]{algorithm2e}

\newcommand*{\affaddr}[1]{#1} 
\newcommand*{\affmark}[1][*]{\textsuperscript{#1}}

\newenvironment{conditions}
{\par\vspace{\abovedisplayskip}\noindent
	\begin{tabular}{>{$}l<{$} @{} >{${}}c<{{}$} @{} l}}
	{\end{tabular}\par\vspace{\belowdisplayskip}}

\newenvironment{conditions*}
{\par\vspace{\abovedisplayskip}\noindent
	\tabularx{\columnwidth}{>{$}l<{$} @{}>{${}}c<{{}$}@{} >{\raggedright\arraybackslash}X}}
{\endtabularx\par\vspace{\belowdisplayskip}}

\journalname{}
\begin{document}
\sloppy

\title{A High-Performance CNN Method for Offline Handwritten Chinese Character Recognition and Visualization
}

\titlerunning{A High-Performance CNN for Offline HCCR}   

\author{Pavlo Melnyk\protect\affmark[1,2]  \and Zhiqiang You\protect\affmark[1,2] \and Keqin Li\protect\affmark[2,3] 
}

\authorrunning{P. Melnyk et al.} 

\institute{
	\Letter\tabto*{0.55cm}Z. You \at
	\tabto*{0.55cm}{you@hnu.edu.cn}   
	\and
	\tabto*{0.55cm}P. Melnyk \at
	\tabto*{0.55cm}{pavlomelnyk@hnu.edu.cn}
	\and 
	\tabto*{0.55cm}K. Li \at
	\tabto*{0.55cm}{lik@newpaltz.edu}
	\and\\ 
	\affaddr{\affmark[1]\tabto*{0.55cm}Key Laboratory for Embedded and Network Computing \\
	\tabto*{0.55cm}of Hunan Province, Hunan University, Changsha 410082, \\
	\tabto*{0.55cm}People’s Republic of China}\\\\     
	\affaddr{\affmark[2]\tabto*{0.55cm}College of Computer Science and Electronic Engineering, \\
	\tabto*{0.55cm}Hunan University, Changsha 410082, People’s Republic \\
	\tabto*{0.55cm}of China}\\\\
	\affaddr{\affmark[3]\tabto*{0.55cm}Department of Computer Science, State University of \\
	\tabto*{0.55cm}New York at New Paltz, New Paltz, NY 12561, USA}        
}

\date{Received: date / Accepted: date}

\maketitle

\begin{abstract}
Recent researches introduced fast, compact and efficient convolutional neural networks (CNNs) for offline handwritten Chinese character recognition (HCCR). However, many of them did not address the problem of network interpretability. We propose a new architecture of a deep CNN with high recognition performance which is capable of learning deep features for visualization. 
A special characteristic of our model is the bottleneck layers which enable us to retain its expressiveness while reducing the number of multiply-accumulate operations and the required storage. We introduce a modification of global weighted average pooling (GWAP) - global weighted output average pooling (GWOAP). 
This paper demonstrates how they allow us to calculate class activation maps (CAMs) in order to indicate the most relevant input character image regions used by our CNN to identify a certain class. Evaluating on the ICDAR-2013 offline HCCR competition dataset, we show that our model enables a relative 0.83\% error reduction while having 49\% fewer parameters and the same computational cost compared to the current state-of-the-art single-network method trained only on handwritten data. Our solution outperforms even recent residual learning approaches. 
\end{abstract}

\section{Introduction}
\label{intro}

\begin{figure*}
	\centering
	\makebox[\textwidth]{\includegraphics[width=\textwidth]{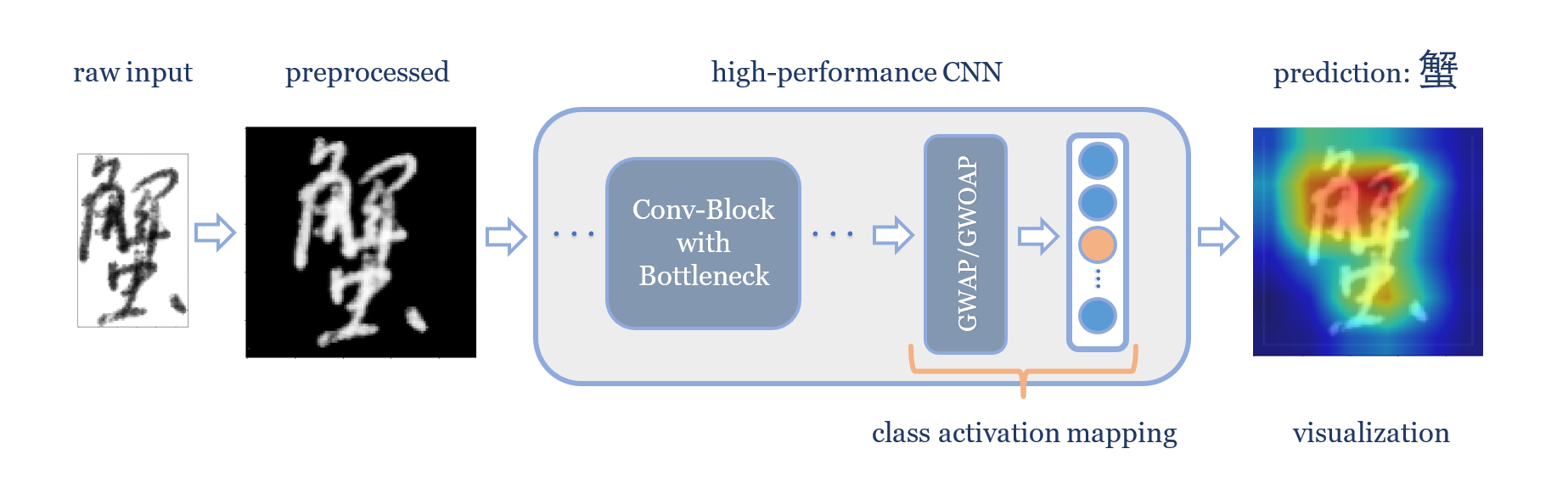}}
	\caption{A high-performance method for offline HCCR and visualization by means of CAMs}
	\label{fig:0}
\end{figure*}

With the rapid development of deep learning technologies, many tasks regarding pattern recognition have obtained considerable improvements. The tasks vary significantly from object detection and image generation to spinning articles and generating poetry. The problem of text recognition is also a good example of learning discriminative representations performed by deep learning algorithms. 

Text recognition at the character level can be divided into printed and handwritten character recognition. Automatic recognition of medical forms and processing of other types of files, such as administrative, postal mail sorting automation and bank checks identification, are all examples of applications for handwritten character recognition which may further be either offline or online.

In this regard, the problem of offline handwritten Chinese character recognition (HCCR) \citep{liu2013online, li2016handwritten} having been studied for more than half a century is of particular interest. Offline HCCR involves analysis and classification of isolated handwritten Chinese character images. 
Earlier successful methods for offline HCCR, such as modified quadratic discriminant functions (MDQF) \citep{kimura1987modified}, were effectively and significantly outperformed by convolutional neural network (CNN) approaches. Noteworthy, these days some hybrid methods, such as those utilizing adversarial feature learning \citep{zhang2018robust} or an attention-based recurrent neural network (RNN) for iterative refinement of the predictions \citep{yang2017improving}, seem to be the next effective substitution for the traditional CNN solutions \citep{cirecsan2012multi,yin2013icdar,zhong2015high,zhang2017online}. 

However, in our study, we create a method based on a pure CNN architecture, with high recognition performance while keeping in mind its size and computational cost.
Notwithstanding that different data augmentation as well as feature handcrafting and spatial-transforming techniques were successfully utilized for offline HCCR, we refrain from using such in order to focus our work mainly on finding optimal hyperparameters for training only on the raw handwritten input data. 
	The reasons why we choose the approach with deep neural networks are their high-recognition performance for large-scale classification tasks, the availability of end-to-end training, and the ongoing research on their improvements.

One of the main reasons causing shortcomings of the CNNs is the network interpretability \citep{qin2018convolutional}. This question is especially interesting for us in terms of such a large-scale classification problem as offline HCCR. In this domain, both low-level visual features, such as small strokes and their high-level structural concatenations, are important for making correct predictions \citep{yang2017improving}.

In order to address this issue, we adopt the knowledge of class activation maps (CAMs) \citep{zhou2016learning}. We demonstrate how it improves the network interpretability by performing a visualization of the most relevant character parts learned by it. Unlike the visualization of the network layer outputs as was done in the context of offline HCCR by \cite{zhang2015deep}, exploiting CAMs allows understanding the process from the beginning to the end. 

The main contributions of our work are 
	summarized in Fig.~\ref{fig:0}
:
1) we propose a CNN model for offline HCCR, which achieves state-of-the-art accuracy for single-network methods trained only on handwritten data; 
2) we employ modified versions of global average pooling (GAP) - global weighted average pooling (GWAP) and the introduced global weighted output average pooling (GWOAP) to obtain high performance and accomplish the visualization.

The rest of the paper is organized as follows:
Sect.~\ref{sec:1} reviews related research; Sect.~\ref{sec:4} describes the proposed architecture and its effectiveness, introduces our modification of GWAP - GWOAP, and also details how CAMs can be computed when a network is equipped with either; Sect.~\ref{sec:8} shares the implementation details and the results of our experiments including a comparison with other methods; and Sect.~\ref{sec:16} summarizes our work.


\section{Related Work}
\label{sec:1}
\subsection{Offline HCCR}
\label{sec:2}
The reasons why HCCR is a non-trivial problem can be mainly formulated as follows:

1) writing variations;

2) wide-scale vocabulary - the number of character classes ranges from 6763 to 70244 in GB2312-80 and GB18010-2005 standards, respectively;

3) similarities between Chinese characters.

Nowadays, achievements in deep learning enable researchers to successfully utilize CNNs in the HCCR domain \citep{zhong2015high,cheng2016handwritten,zhong2016handwritten,xiao2017building,Li2018}, greatly outperforming MDQF methods \citep{kimura1987modified,lu2015cost}. Such a CNN was first applied to this problem by \cite{cirecsan2012multi}. Their single multi-column deep neural network (MCDNN) achieves a 94.47\% accuracy. 

Later, works were evaluated on the ICDAR-2013 competition \citep{yin2013icdar} dataset containing 3755  character classes, which corresponds to the key official character set GB2312-80 level-1. 

There is a very noticeable trend in the offline HCCR competition: the better deep CNNs perform, the more different aspects researchers consider for their models. For instance, the Fujitsu research team created a CNN-based method and took the winner place in the ICDAR-2013 competition with an accuracy of 94.77\% \citep{yin2013icdar} while requiring as much as 2460MB for storage.

The first model that outperformed human-level performance was introduced by \cite{zhong2015high}, which incorporated traditional directional feature maps. Therefore, their single HCCR-Gabor-GoogLeNet and ensemble HCCR-Ensemble-GoogLeNet-10 models achieve a recognition accuracy of 96.35\% and 96.74\% and have a size of 27.77MB and 270.0MB, respectively.

\cite{cheng2016handwritten} showed how the combination of the character classification and similarity ranking supervisory signals increases inter-class variations and reduces intra-class variations. Their single deep CNN  achieves a 97.07\% accuracy taking 36.80MB for storage. The ensemble of four such networks has a better performance of 97.64\%.

\cite{zhong2016handwritten} introduced a network composed of two parts: a spatial transformer network rectifying the input image and a deep residual network predicting the label distribution for the rectified image, which resulted in an accuracy of 97.37\% with a 92.30MB storage required. 

\cite{zhang2017online} used the traditional normalization-cooperated direction-decomposed feature map (DirectMap) along with the deep CNNs to obtain an accuracy of 96.95\%, further improved to 97.37\% by introducing adaptation layer aimed at reducing the mismatch between training and test samples on a particular source layer. Both models have a size of 23.50MB. It takes 1.997ms to calculate DirectMap and 296.894ms to perform a forward pass of a deep CNN for processing a character image on a CPU. 

A method using residual blocks \citep{he2016deep} and iterative model prediction refinement by means of an attention-based RNN is a hybrid approach proposed by \cite{yang2017improving}. They achieved an accuracy of 97.37\%, outperforming previous methods that used raw input data.

A fast and compact CNN was developed by \cite{xiao2017building} with a speed of 9.7ms/char on a CPU but only 2.3MB of storage needed and an accuracy of 97.09\%. That was enabled by employing global supervised low-rank expansion (GLSRE) and adaptive drop-weight techniques (ADW). In their experiments, one of the baseline models with a size of 48.7MB  yielded a state-of-the-art accuracy of 97.59\%, considering single-network methods trained only on handwritten data.

Another well-balanced network in terms of the speed, size, and performance was recently introduced by \cite{Li2018}. Their cascaded single-CNN model takes only 6.93ms to classify a character image on a CPU, and achieves an accuracy of 97.11\% requiring only 3.3MB for storage. They accomplished this by utilizing fire-modules and the proposed GWAP concept along with quantization. 

One of the newest methods reported by \cite{zhang2018robust} introduced adversarial feature learning (AFL), which significantly outperforms traditional deep CNN approaches by exploiting writer-independent semantic features with the prior knowledge of standard printed characters, resulting in a 98.29\% test set accuracy and an 18.2MB model size.

	A new family of DNNs, namely ordinary differential equations (ODEs), were recently proposed by \cite{chen2018neural}. They can successfully be used as supervised learning and time-series models. This could very well be a prospective improvement for the existent HCCR methods. However, at the time of writing, such approaches were not introduced.

\subsection{Class Activation Maps}
\label{sec:3}
\cite{zhou2016learning} presented a method of generating CAMs showing how GAP proposed by \cite{lin2013network} enables a CNN trained for the object recognition task to perform object localization. This technique allows indicating the most important for classification regions of an input image. The main idea behind lies in basic knowledge of a CNN structure: as we move deeper, the height and width of feature maps shrink, while the number of channels increases. 

GAP used instead of the traditional fully connected layer at the end of the network produces the spatial average of every channel of the preceding convolution layer output. Later, the weighted sum of these values is used in order to generate final output – perform a logistic regression. Remarkably, it is easily interpretable: one can think of a feature going into the logistic regression as a value indicating whether or not something important for classification appears in the image. 

Similarly, a CAM is a weighted sum of the GAP input (the last convolution layer output), i.e., if we were to look at the image before the spatial averaging, we would know where exactly a distinctive region was. It is worth mentioning that we consider only one class when producing a CAM – the predicted class. 

Let the output of the last convolution layer be a 3-D tensor $\textbf{F} \in \mathbb{R} {^{H{\rm{ }} \times {\rm{ }}W{\rm{ }} \times {\rm{ }}C}}$, the output of the GAP be a vector ${\textbf{f}_{out}} \in \mathbb{R} {^C}$, and the logistic regression weight matrix be ${\textbf{W}_{out}} \in \mathbb{R} {^{C{\rm{ }} \times {\rm{ }}K}}$. In order to calculate the activation map, all one needs to do is weigh the importance of each $H \times W$  feature of $\textbf{F}$  by multiplying them by the corresponding elements of the column of ${\textbf{W}_{out}}$ that connects $\textbf{f}_{out}$ to the predicted class output:
\begin{equation}
\label{eq:cam}
\textbf{CAM} = \sum\limits_{i = 1}^C {{w_{out}}_{i}^{k'}} \cdot  {\textbf{F}_i}~,
\end{equation}
where: 
\begin{conditions}
	H, W, C & - & height, width and number of channels of the\\
	&& feature map,\\
	K & - & total number of classes,\\   
	k' & - & the predicted class.
\end{conditions}

One can notice that (\ref{eq:cam}) is a dot-product between the $k'$-th  weight vector of the matrix $\textbf{W}_{out}$  and the last conv-layer output feature map $\textbf{F}$. 
We can simply zoom the obtained CAM to the size of the input image, thus identify the image regions most relevant to the certain category. 

Notably, this strategy for visualization is different from the one exploited by \cite{yang2017improving}. In their multi-scale residual block cascade, they introduced shortcut connections that aggregate “lower” and “higher” layer activations with different heights and widths but the same number of channels and defined an aggregation operation as the union of feature vectors. The obtained in this way learned visual representation was proposed to be fed into the iterative refinement module to improve the classification performance.
\begin{figure*}
	\centering
	\makebox[\textwidth]{\includegraphics[width=\textwidth]{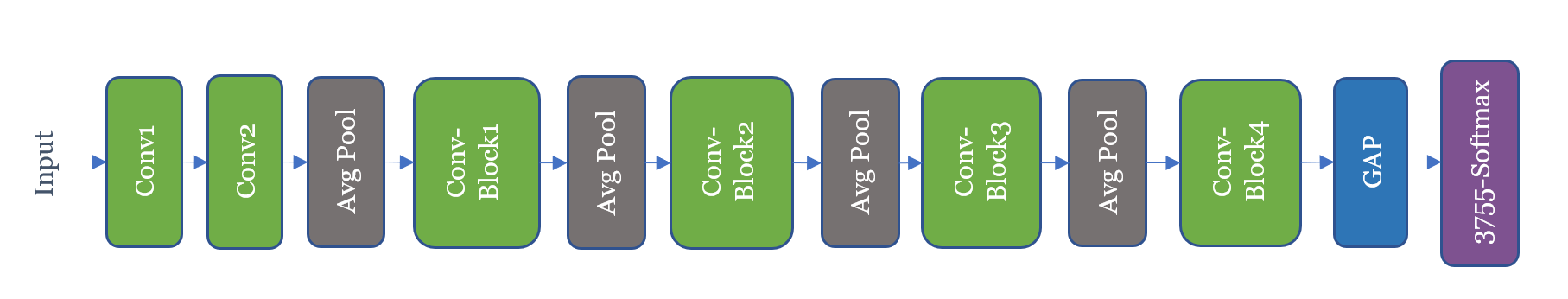}}
	\caption{Baseline architecture of the proposed network}
	\label{fig:1}
\end{figure*}

\section{Method Description}
\label{sec:4}

\subsection{Proposed Architecture}
\label{sec:5}

Being inspired by the performance gain enabled by utilizing the modification of GAP - GWAP proposed by \cite{Li2018}, we employ it and compare its performance with GAP and our modification – GWOAP. Thus, we could see how good a single deep CNN without residual connections can perform for the offline HCCR task and visualize the most distinctive regions of an input character image. The corresponding networks are further referred to as Model A, Model B and Model C as shown in Table~\ref{tab:1}. The description of GWAP and the proposed GWOAP is presented in Sect.~\ref{sec:7}.

Similar to the state-of-the-art method \citep{xiao2017building}, we resize input images to $96 \times 96$, as smaller samples result in a poorer accuracy, while bigger – in an expensive computational cost. Every conv-layer in our network has kernels of a $3 \times 3$ size, a stride parameter of 2, and retains spatial dimensionality of input – performs the “same” mode of padding. Batch normalization (BN)  \citep{ioffe2015batch} is a technique that was used as a default choice by many researchers over the past few years and has been proven to be effective for offline HCCR \citep{xiao2017building}. Therefore, we equip every conv-layer with a BN-layer followed by a rectified linear unit (ReLU). Importantly, we do not use biases for the conv-layers, because it is redundant due to the presence of the second location parameters $\beta$  in BN-layers.

The proposed baseline architecture (Model A) is presented in Fig.~\ref{fig:1}. It consists of 15 layers, counting only convolutional and fully-connected ones. We use average pooling layers with $3 \times 3$ windows and a stride parameter of 2.
The easiest way to describe our model is in terms of convolutional blocks – groups of three convolutional layers with a \textit{bottleneck} in the middle. The hyperparameters are shared between the three conv-layers, except for the number of kernels in the bottleneck. We discuss the effectiveness of such conv-blocks in the next subsection. 

The first two conv-layers in our model are followed by a pooling layer, and then by 4 conv-blocks separated by pooling layers. Final conv-block produces a feature map of a $6 \times 6 \times 448$ size. Later, it is fed into GAP which outputs a vector of a 448 length. It is further connected to a 3755-softmax output, where the number of units corresponds to the number of character classes considered in this work.

Remarkably, smaller sizes of the last conv-layer output result in more blurry CAMs, as we need to upsample the maps to the size of the input. Through experiments, we observe that $6 \times 6$ output feature maps and $96 \times 96$  input images represent a well-balanced trade-off between the model performance and the visual clarity of the obtained CAMs.

\newcommand{\tabincell}[2]{\begin{tabular}{@{}#1@{}}#2\end{tabular}}%
\begin{table*} 
	\tabcolsep=7pt{
		\centering
	\renewcommand\arraystretch{1.5}
		\caption{The three networks proposed for offline HCCR}
		\label{tab:1}%
		\begin{tabular}{ccccc}
			\toprule
			\tabincell{l}{\textbf{Layer} \\\textbf{Name}} & 
			\multicolumn{1}{c}{\textbf{Model A}} & \multicolumn{1}{c}{\textbf{Model B}} & \multicolumn{1}{c}{\textbf{Model C}}
			& \tabincell{l}{\textbf{Output} \\ \textbf{Shape}}\\
			\midrule
			Input & &\multicolumn{1}{l}{96 x 96 grayscale image}& & \multicolumn{1}{l}{96 x 96 x 1} \\
			\hline
			Conv1 & &\multicolumn{1}{l}{3 x 3 conv. 64, BN, ReLU} && \multicolumn{1}{l}{96 x 96 x 64} \\
			\hline
			Conv2 & &\multicolumn{1}{l}{3 x 3 conv. 64, BN, ReLU} && \multicolumn{1}{l}{96 x 96 x 64} \\
			\hline
			AvgPool & &\multicolumn{1}{l}{3 x 3 avg-pool stride 2} && \multicolumn{1}{l}{48 x 48 x 64} \\
			\hline
			\multirow{3}[2]{*}{\tabincell{l}{Conv-Block1}} & &\multicolumn{1}{l}{3 x 3  conv. 96, BN, ReLU} & & \multicolumn{1}{l}{48 x 48 x 96} \\
			\multicolumn{1}{l}{} & &\multicolumn{1}{l}{3 x 3 conv. 64, BN, ReLU} && \multicolumn{1}{l}{48 x 48 x 64} \\
			\multicolumn{1}{l}{} & &\multicolumn{1}{l}{3 x 3 conv. 96, BN, ReLU} && \multicolumn{1}{l}{48 x 48 x 96} \\
			\hline
			AvgPool & &\multicolumn{1}{l}{3 x 3  avg-pool stride 2}& & \multicolumn{1}{l}{24 x 24 x 96} \\
			\hline
			\multirow{3}[2]{*}{\tabincell{l}{Conv-Block2}} & &\multicolumn{1}{l}{3 x 3  conv. 128, BN, ReLU} && \multicolumn{1}{l}{24 x 24 x 128} \\
			\multicolumn{1}{l}{} & &\multicolumn{1}{l}{3 x 3  conv. 96, BN, ReLU} && \multicolumn{1}{l}{24 x 24 x 96} \\
			\multicolumn{1}{l}{} & &\multicolumn{1}{l}{3 x 3  conv. 128, BN, ReLU}& & \multicolumn{1}{l}{24 x 24 x 128} \\
			\hline
			AvgPool  & & \multicolumn{1}{l}{3 x 3  avg-pool stride 2} && \multicolumn{1}{l}{12 x 12 x 128} \\
			\hline
			\multirow{3}[2]{*}{\tabincell{l}{Conv-Block3}} & &\multicolumn{1}{l}{3 x 3  conv. 256, BN, ReLU} && \multicolumn{1}{l}{12 x 12 x 256} \\
			\multicolumn{1}{l}{} & &\multicolumn{1}{l}{3 x 3  conv. 192, BN, ReLU} && \multicolumn{1}{l}{12 x 12 x 192} \\
			\multicolumn{1}{l}{} & &\multicolumn{1}{l}{3 x 3  conv. 256, BN, ReLU} && \multicolumn{1}{l}{12 x 12 x 256} \\
			\hline
			AvgPool & &\multicolumn{1}{l}{3 x 3  avg-pool stride 2} && \multicolumn{1}{l}{6 x 6 x 256} \\
			\hline
			\multirow{3}[2]{*}{\tabincell{l}{Conv-Block4}} & &\multicolumn{1}{l}{3 x 3  conv. 448, BN, ReLU} && \multicolumn{1}{l}{6 x 6 x 448} \\&& \multicolumn{1}{l}{3 x 3  conv. 256, BN, ReLU} && \multicolumn{1}{l}{6 x 6 x 256} \\&&
			\multicolumn{1}{l}{3 x 3  conv. 448, BN, ReLU} && \multicolumn{1}{l}{6 x 6 x 448} \\
			\hline
			\tabincell{l}{GAP/\\ GWOAP/ \\GWAP} & \multicolumn{1}{l}
			{ \tabincell{l|}{global avg-pool over\\ $6 \times 6$ spatial dims.}} & {\tabincell{c}{global weighted output avg-pool \\ over $6 \times 6$ spatial dims.}} & {\tabincell{|l}{global weighted avg-pool\\ over $6 \times 6$  spatial  dims.}} & 448 \\
			\hline
			Output & \multicolumn{3}{c}{3755-way Softmax} & 3755 \\
			\bottomrule
		\end{tabular}%
	}
\end{table*}%

\subsection{Effectiveness of the Convolutional Block Bottleneck}
\label{sec:6}
The effectiveness of bottleneck layers in the proposed model for offline HCCR is proven empirically: they allow the network to retain its expressiveness while reducing the number of multiply-accumulate operations and the required storage.

Considering a single conv-layer, let $H$ and $W$ be the size of the input feature map, $C$ be the number of channels per input feature map, $R$ and $Q$ be the size of the kernel of the conv-layer, $M$, $P$, $S$ be the number of kernels, the size of zero-padding, and the stride, respectively. Then the number of multiply-accumulations (MAC) $N_{MAC}$ for the conv-layer can be calculated as follows:
\begin{equation}
\label{eq:num_mac}
\begin{split}
{N_{MAC}} \!=\! &~ \frac{{H + 2P - R + S}}{S} \cdot \frac{{W + 2P - Q + S}}{S}\\ &~~~~ \cdot R \cdot Q \cdot C \cdot M~.
\end{split}
\end{equation}

Assuming that all conv-layers retain the dimensionality of input ($H \times W$), the total number of MAC in this conv-block can be found in accordance with (\ref{eq:num_mac}) as:
\begin{equation}
\begin{split}
N_{MAC\_CB} = &~ \frac{{H + 2P - R + S}}{S}\! \cdot \! \frac{{W + 2P - Q + S}}{S}\\ &~~~~ \cdot R \cdot Q \cdot C \cdot M  \\ &~ +2 \! \cdot \! \frac{{H + 2P - R + S}}{S} \! \cdot \! \frac{{W + 2P - Q + S}}{S}\\ &~~~~ \cdot R \cdot Q \cdot M \cdot M~.
\end{split}
\end{equation}

Let the middle layer of the conv-block be a bottleneck that outputs a $H \times W \times M_B$ volume. Then the number of MAC in such a conv-block can be calculated as:
\begin{equation}
\begin{split}
N_{MAC\_C{B^{^*}}} = &~
 \frac{{H + 2P - R + S}}{S} \! \cdot \! \frac{{W + 2P - Q + S}}{S} \\ &~~~~\cdot R \cdot Q \cdot C \cdot M \\
 &~  +2 \! \cdot \! \frac{{H + 2P - R + S}}{S}\! \cdot \!\frac{{W + 2P - Q + S}}{S} \\ &~~~~\cdot R \cdot Q \cdot M_B \cdot M~.
\end{split}
\end{equation}

Thus, the reduction in computation as well as in storage can be found as:
\begin{equation}
\frac {N_{MAC\_CB}}{N_{MAC\_C{B^{^*}}}} = \frac{C + {\rm{2}}M}{C + {\rm{2}}{M_B}}~.
\end{equation}

On the other hand, utilizing bottlenecks can be viewed as a compression-decompression operation, which is kind of regularization itself.

\subsection{Obtaining Class Activation Maps with GWAP and GWOAP}
\label{sec:7}
GAP \citep{lin2013network} performs spatial averaging, i.e.:
\begin{equation}
\begin{aligned}
GAP(\textbf{F}) =&~ \frac{1}{HW}\sum\limits_{i = 1}^H {\sum\limits_{j = 1}^W {f_{ijm}}},~~1 \le m \le C~.
\end{aligned}
\end{equation}

Its modification, GWAP \citep{Li2018}, can be simply expressed as:
\begin{equation}
\begin{aligned}
GWAP(\textbf{F}) =&~ \sum\limits_{i = 1}^H {\sum\limits_{j = 1}^W {w_{ijm}^{GWAP} \cdot {f_{ijm}}} },~~1 \le m \le C~.
\end{aligned}
\end{equation}

The proposed modification of GAP is defined as:
\begin{equation}
\begin{aligned}
GWOAP(\textbf{F}) =&~ \sum\limits_{i = 1}^H {\sum\limits_{j = 1}^W{w_{m}^{GWOAP} \cdot f_{ijm}}},~~1 \le m \le C~,
\end{aligned}
\end{equation}
where:
\begin{conditions*}
\textbf{F} \in {\mathbb{R}^{H \times W \times C}} & - & input feature map,\\
\textbf{W}_{}^{GWAP} \in {\mathbb{R}^{H \times W \times C}} & - & 3-D trainable kernel of GWAP,\\
\textbf{w}_{}^{GWOAP} \in {\mathbb{R}^C} & - & 1-D trainable kernel of GWOAP.
\end{conditions*}

The difference between the two modifications is the number of parameters: GWOAP scales the output of spatial summation, rather than its input, which is more in the “convolutional” manner, i.e., sharing learnable scaling parameters channel-wise rather than shape-wise. It also can be seen as regularization.
The process of obtaining CAMs for the network equipped with either GWAP or GWOAP is different from the one for the network equipped with GAP (\ref{eq:cam}) only by one additional operation (\ref{eq:scale_cam}). The output of the last conv-layer $\textbf{F}$ is to be scaled by the trainable kernel -  either $\textbf{W}^{GWAP}$ or $\textbf{w}^{GWOAP}$:
\begin{equation}
\label{eq:scale_cam}
{\textbf{F}^*} = \left\{ \begin{gathered}
\textbf{W}_{}^{GWAP} \odot \textbf{F} \hfill \\
\textbf{W}_{}^{GWOAP} \odot \textbf{F} \hfill\\
\end{gathered}  \right.~,
\end{equation}
\begin{equation}
\textbf{CAM} = \sum\limits_{i = 1}^C {{w_{out}}_i^{k'} \cdot \textbf{F}_i^*}~,
\end{equation}
where:
\begin{conditions*}
\textbf{W}_{}^{GWOAP} & - & upsampled 3-D version of $\textbf{w}_{}^{GWOAP}$\\ &&for performing a valid element-wise multiplication $\odot$,\\
\textbf{F}^* & - & 3-D scaled feature map,\\
\textbf{CAM} & - & obtained 2-D class activation map.
\end{conditions*}

	The calculation of CAMs is described in Algorithm~\ref{alg:cam}.
We compare how well both modifications of GAP perform in offline HCCR competition and discuss the CAMs produced by means of the proposed models for different character images in Sect.~\ref{sec:14}.

\IncMargin{1em}

\begin{algorithm}
	\caption{Class activation mapping for offline HCCR with the proposed models}\label{alg:cam}
	\SetKwInOut{Input}{input}\SetKwInOut{Output}{output}
	\Input{A character $Im$ of size $96\times 96$}
	\Output{A class activation map $Im$ of size $96\times 96$}
	\BlankLine
	\textbf{Step 1}\\
	$character\_class = model.predict(input)$\\ \tcp{perform a forward pass through the model and get the prediction}
	\BlankLine
	\textbf{Step 2}\\
	$fmaps = model.get\_output(last\_conv\_layer,~input)$\\ \tcp{get the output of the last conv-layer}
	\BlankLine
	\textbf{Step 3}\\
	\If{$GWOAP$ {\bf or} $GWAP$}{
		$fmaps = fmaps \odot W^{GWOAP/GWAP}$\\
	\tcp{scale feature maps with the GWOAP/GWAP kernel}}
	\BlankLine
	\textbf{Step 4}\\
	$W_{out} = model.output\_layer.get\_weights()$\\
	${w_{out}} = W_{out}[:,~character\_class]$\\
	\tcp{get the last layer weights connecting GAP/GWOAP/GWAP output to the predicted class}
	\BlankLine
	\textbf{Step 5}\\
	$CAM = fmaps.dot(w_{out})$\\
	$CAM = upsample(CAM,~size=(96, 96),\newline ~~~~~~~~~~~~~~~~~~~~~~~~~interpolation="bilinear")$\\
	\tcp{compute CAM and upsample it to the size of input using bilinear interpolation}	
	\BlankLine	
	\textbf{return} $CAM$
	\BlankLine
\end{algorithm}\DecMargin{1em}

\section{Experiments}
\label{sec:8}
In this section, we share the implementation details and demonstrate the effectiveness of the proposed models not only in terms of recognition performance, but also from the visualization perspective.
\subsection{Datasets}
\label{sec:9}
In order to train the proposed networks, we use CASIA-HWDB1.0-1.1 \citep{liu2011casia} datasets collected by National Laboratory of Pattern Recognition (NLPR), Institute of Automation of Chinese Academy of Sciences (CASIA), written by 420 and 300 persons, respectively.

The overall training dataset contains 2,678,424 samples belonging to 3755 different character classes. We evaluate our models on the most common benchmark for offline HCCR – the ICDAR-2013 competition dataset \citep{yin2013icdar}, containing 224,419 samples written by 60 persons.

It is worth mentioning that we do not use the test set as validation data for finding hyperparameters. The validation set of a 60,000 samples size is randomly selected from the training data. After finding optimal settings for our models, we merge validation and training sets and conduct further experiments.

We use raw images: the only data preprocessing we make is normalization to $96 \times 96$ size and inversion of the pixel intensity.

\subsection{Training Strategy}
\label{sec:10}
First, we shuffle the training data. The parameters of all conv-layers are set with the He-normal initialization \citep{he2015delving}. The classification layer weights are randomly initialized by drawing from a Gaussian distribution with the standard deviation of 0.001, while the bias term is initially set to 0. The parameters of GWAP and GWOAP are initialized with 1.

We use stochastic gradient descent (SGD) with the momentum term of 0.9 for training, which is a common choice for pretty much every CNN proposed for this competition over the past few years. The mini-batch size is set to be 256, and the maximum number of epochs is 40.

Exploiting batch normalization allows us to choose a higher learning rate value. Initially, we set the learning rate to 0.1 and train our models for one epoch. Then we decrease it by a factor of 10 and keep decreasing after every epoch when training accuracy stops improving. 

To deal with overfitting for all models in our work, we use L2-regularization with the multiplier equal to 0.001, and a dropout \citep{srivastava2014dropout} before the softmax layer, where the probability of dropping, $p_{drop}$, is set to 0.5. Also, we do not use any data augmentation method for generating distorted images during training.

We implement the proposed CNNs using the amazing Keras deep learning library \citep{chollet2015keras} with the TensorFlow \citep{abadi2016tensorflow} backend, and conduct all experiments on NVIDIA GeForce GTX 1080 Ti with 11GB of memory. A single experiment takes 2 days on average.

\subsection{Results}
\label{sec:13}

The performances of the proposed models are demonstrated in Table~\ref{tab:2}.
\begin{table}
	\caption{Comparison of the proposed networks performance. The last column shows the absolute difference between a trained model accuracy and its accuracy when the input bias term for the softmax layer is reset to 0}
	\label{tab:2}%
	\tabcolsep=2.8pt
	\begin{tabular}{p{3.955em}ccc}
		\hline\noalign{\smallskip}
		\textbf{Model} & \multicolumn{1}{p{6.335em}}{\textbf{Parameters}} & \multicolumn{1}{p{5em}}{\textbf{Accuracy}} & \multicolumn{1}{p{9em}}{\textbf{Accuracy Drop}} \\
		\noalign{\smallskip}\hline\noalign{\smallskip}
		A     & 6,507,691 & 97.38\% & 0.00045\% \\
		B     & 6,508,139 & 97.55\% & 0.00045\% \\
		C     & 6,523,819 & 97.61\% & 0.00045\% \\
		\noalign{\smallskip}\hline
	\end{tabular}%
\end{table}%

Obtaining CAMs with Model A is pretty straightforward, since it is equipped with GAP, and can be done in accordance with (1). As for Model B and Model C, we use the additional operation (\ref{eq:scale_cam}). 
Similar to the original source \citep{zhou2016learning}, we ignore the bias of the softmax, 
	when computing CAMs,
as it has little to no impact on the classification accuracy as shown in the rightmost column in Table~\ref{tab:2}.
After that, we upsample the obtained $6 \times 6$ maps to the input image size 
	using bilinear upsampling
and plot them together with the input to visualize its most relevant regions.
The CAMs produced by means of each model are shown in Fig.~\ref{fig:2}. The first two rows display one of the most confusing handwritten Chinese characters pairs – 
\begin{CJK*}{UTF8}{gbsn}
	"已" (yi) and "己" (ji). 
\end{CJK*}
The last two rows show the most distinctive parts of \begin{CJK*}{UTF8}{gbsn}
"鲍" (bao) and "吗" (ma).
\end{CJK*}

\begin{table*}
	\centering
	\caption{Comparison of the ICDAR-2013 offline HCCR competition methods. “Raw Data” shows whether manually preprocessed or raw input images are used. All displayed methods except for Human Level Performance include CASIA HWDB1.0-1.1 in training datasets}
	\label{tab:3}%
	\footnotesize\addtolength{\tabcolsep}{0pt}
	\resizebox{!}{4.8cm}{\begin{tabular}{ccccccc}
			\toprule
			\textbf{Method} & \textbf{Size~(MB)} & \textbf{Accuracy} & \textbf{Ensemble} & \textbf{Raw Data} & {\textbf{Ref.}} \\
			\midrule
			Human Level Performance & n/a   & 96.13\% & n/a   & n/a   & \cite{yin2013icdar} \\
			\midrule
			HCCR-Gabor-GoogLeNet & 27.7  & 96.35\% & no    & yes   & \cite{zhong2015high} \\
			HCCR-GoogLeNet-Ensemble-10 & 270.0 & 96.74\% & yes (10) & yes   &  \\
			\midrule
			Residual-34 & 92.2  & 97.36\% & no    & yes   & \cite{zhong2016handwritten} \\
			STN-Residual-34 & 92.3  & 97.37\% & no    & yes   &  \\
			\midrule
			DCNN-Similarity ranking & 36.2  & 97.07\% & no    & yes   & \textcolor{black}{\cite{cheng2016handwritten}} \\
			\textcolor{black}{Ensemble DCNN-Similarity ranking}   & \textcolor{black}{144.8}  & \textcolor{black}{97.64\%} & \textcolor{black}{yes (4)}   & \textcolor{black}{yes}   & \textcolor{black}{} \\
			\midrule
			DirectMap + ConvNet & 23.5  & 96.95\% & no    & no    &  \\
			DirectMap + ConvNet + Ensemble-3 & 70.5  & 97.12\% & yes (3) & no    & \cite{zhang2017online} \\
			DirectMap + ConvNet + Adaptation & 23.5  & 97.37\% & no    & no    &  \\
			\midrule
			M-RBC + IR & n/a   & 97.37\% & no    & yes   & \cite{yang2017improving} \\
			\midrule
			HCCR-CNN9Layer+GSLRE 4X +ADW & 2.3   & 97.09\% & no    & yes   &  \\
			HCCR-CNN12Layer+GSLRE 4X+ADW & 3.0     & 97.40\% & no    & yes   & \textcolor{blue}{\cite{xiao2017building}} \\
			\textcolor{blue}{HCCR-CNN12Layer} & \textcolor{blue}{48.7}  & \textcolor{blue}{97.59\%}& \textcolor{blue}{no}   & 	\textcolor{blue}{yes}  &   \\			
			\midrule
			Cascaded Model (Quantization) & 3.3   & 97.11\% & no    & yes   &  \cite{Li2018}\\
			Cascaded Model & 20.4  & 97.14\% & no    & yes   &  \\				
			\midrule
			\textcolor{black}{AFL}   & \textcolor{black}{18.2}  & \textcolor{black}{98.29\%} & \textcolor{black}{no}   & \textcolor{black}{yes}   & \textcolor{black}{\cite{zhang2018robust}} \\
			\midrule
			\midrule
			Model A & 24.8  & 97.38\% & no    & yes   &  \\
			Model B & 24.8  & 97.55\% & no    & yes   & ours \\
			\textbf{Melnyk-Net (Model C)} & \textbf{24.9} & \textbf{97.61\%} & \textbf{no} & \textbf{yes} &  \\
			\bottomrule
		\end{tabular}%
	}	
\end{table*}%

\subsection{Comparison of the Proposed Models}
\label{sec:14}
\begin{figure}	
	\centering
	\includegraphics[width=0.4\textwidth]{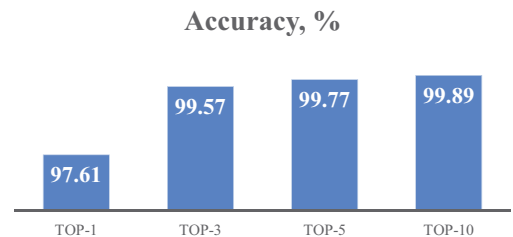}
	\caption{Model~C top-k classification performance on the ICDAR-2013 competition dataset}
	\label{fig:3}  
\end{figure}

Model C utilizing GWAP outperforms the other two. Compared to the baseline model (Model A), its number of parameters is bigger only by less than 0.25\%, while it results in a relative performance gain of 0.24\%. 
	The top-\textit{k} accuracies of Model C are demonstrated in Fig.~\ref{fig:3}. It takes 4.15ms averagely to classify a character image on the GPU.

Despite the fact that GWOAP has 36 times fewer parameters than GWAP, Model B is outperformed by Model C with a rather small margin. Nevertheless, as the result suggests, the more attention parameters for global spatial averaging at the end of the network we use, the better it performs on the unseen data. 

\begin{figure}	
	\includegraphics[width=0.5\textwidth]{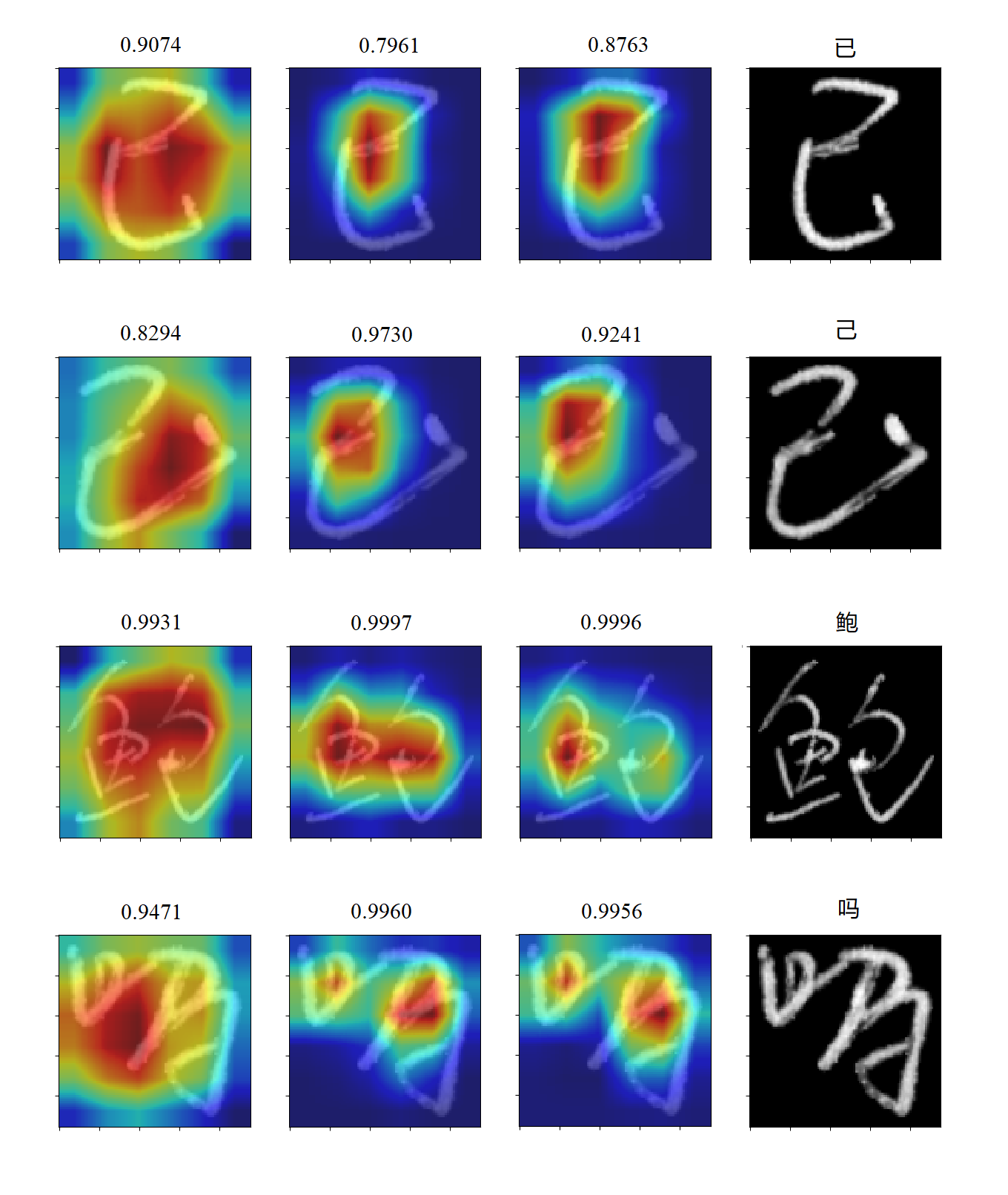}
	\caption{Class activation maps for correctly classified character images and the respective softmax outputs. The first three columns correspond to the class activation maps produced using deep features of Model A, Model B, and Model C, respectively. The last column contains original input images from the ICDAR-2013 competition dataset}
	\label{fig:2}  
\end{figure}

It is very noticeable that not only do Model B and Model C yield the best classification performance, but also by using them, we obtain a comparatively more accurate mechanism for visualization as demonstrated in Fig.~\ref{fig:2}. 

\subsection{Comparison with Other Methods}
\label{sec:15}
The comparison of the ICDAR-2013 offline HCCR competition methods is shown in Table~\ref{tab:3}.

Two of the proposed models, namely Model B and Model C, show a competitive with the state-of-the-art single-CNN method \citep{xiao2017building} classification performance, while it has the same computational cost of 1.2GFLOPs (multiply-accumulations) and is almost twice as small in size. But we do not use a fully connected layer before the classification layer, because it would not allow us to compute CAMs. 

Remarkably, even our baseline network (Model A) of 15 layers of depth outperforms some recent approaches with residual learning \citep{zhong2016handwritten,yang2017improving}. Even though the model proposed by \cite{zhong2016handwritten} uses the spatial transformation of the input images at the network level. It is also worth mentioning that similar to our work, the method described by \cite{yang2017improving} allows visualizing distinctive regions of input character images. However, it utilizes a multi-scale residual block cascade that learns a hierarchy of visual features from the input for iterative refinement of the predictions.

As for the cascaded model \citep{Li2018} which uses GWAP, it is built with the efficiency criteria in mind in order to balance the accuracy, speed and number of parameters and is not aimed to perform the visualization as we do. 

Unlike \cite{cheng2016handwritten}, we do not use data augmentation to generate more samples. All our networks outperform their deep CNN with large margins requiring a 31.5\% smaller storage. However, the ensemble of four such CNNs achieves a 0.03\% relatively higher accuracy being almost 6 times bigger in size compared to the proposed Model C. It is included in our comparison to show a complete picture of the competition.

The AFL method \citep{zhang2018robust} that achieves the state-of-the-art result in offline HCCR competition is hard to compare to our work since their model involves a discriminator guiding the feature extractor to learn the prior knowledge of standard printed characters. On the contrary, we use only handwritten data for training. Additionally, the feature extractor in their network is followed by a fully connected layer, which does not allow to utilize the visualization method exploited in our work.

\subsection{Advantages and Limitations}
Model C performing the best among the proposed three networks is called Melnyk-Net. 
The main advantages of Melnyk-Net are its high recognition performance and size. Even without applying novel comprising techniques, it has almost twice as few parameters and the same computational cost compared to the previous state-of-the-art single-CNN method trained only on handwritten character samples. 
Moreover, our model is equipped with GWAP which allows us to perform class activation mapping in order to indicate the most relevant input character image regions. Thus, we make a step toward interpreting the CNN built for character recognition.
However, the assessment of CAMs in terms of offline HCCR is very subjective since there is no numerical measure for this unlike the object localization task 
(e.g., the Jaccard index).

\section{Conclusion}
\label{sec:16}
In this paper, we propose a high-performance CNN for offline HCCR called Melnyk-Net. 
To the best of our knowledge, it yields state-of-the-art accuracy for single-network methods trained only on handwritten data. Compared to the previous state-of-the-art model, Melnyk-Net is 0.02\% more accurate, while having the same computational cost and requiring almost twice as small storage. We accomplish this by exploiting convolutional layers with bottlenecks and the variation of the global averaging operation. Importantly, Melnyk-Net being 15 layers deep and having no residual connections outperforms recent ResNet-based methods.
Moreover, we show how utilizing GAP and its modifications including the proposed GWOAP enables calculating CAMs in order to perform visualization of the most distinctive regions of an input character image. It improves the network interpretability and can serve as a good tool for classification error analysis in such a large-scale recognition problem as offline HCCR. 

In future work, 
	modern comprising methods can be used to reduce the model size and speed up the computational process. Adding printed data to the training set in conjunction with new samples, e.g., generated by generative adversarial networks (GANs), may be a promising way to further boost the performance.
Also, the visualization obtained by means of CAMs can be exploited in order to learn from and consequently refine the prediction for misclassified samples.
In addition, although increasing the number of character classes may cause classification performance loss, it will broaden the applicability of the offline HCCR model. 
Finally, the newly developed ODE networks must be an advantageous choice for developing much more efficient solutions for the HCCR task.

\begin{acknowledgements}
	This work is supported by National Natural Science Foundation of China under Grant No. 61472123, Hunan Provincial Natural Science Foundation under Grant No. 2018JJ2064. We would like to express our gratitude to the China Scholarship Council for giving the first author an opportunity to obtain Master's degree at Hunan University under Chinese Government Scholarship. 
\end{acknowledgements}

\small
\section*{Compliance with ethical standards}
\label{sec:17}
\textbf{Conflict of interest} All authors declare that they have no conflict of interest regarding the publication of this paper.

\small
\bibliographystyle{spbasic}      

\end{document}